# Deep Reinforcement Learning with a Stage Incentive Mechanism of Dense Reward for Robotic Trajectory Planning

Gang Peng [a,b,c], Jin Yang [a,b,d,*], Xinde Li [a,e,*], Mohammad Omar Khyam [c,f]

[a] Key Laboratory of Image Processing and Intelligent Control, Ministry of Education;

[b] School of Artificial Intelligence and Automation, Huazhong University of Science and Technology, Wuhan, China;

[c] IEEE Member, [d] IEEE Student Member

[e] IEEE senior member, School of Automation, South East University, Nanjing, China

[f] School of Engineering and Technology, Central Queensland University, Melbourne, Australia

*Abstract*----To improve the efficiency of deep reinforcement learning (DRL)-based methods for robot manipulator trajectory planning in random working environments, we present three dense reward functions. These rewards differ from the traditional sparse reward. First, a posture reward function is proposed to speed up the learning process with a more reasonable trajectory by modeling the distance and direction constraints, which can reduce the blindness of exploration. Second, a stride reward function is proposed to improve the stability of the learning process by modeling the distance and movement distance of joint constraints. Finally, in order to further improve learning efficiency, we are inspired by the cognitive process of human behavior and propose a stage incentive mechanism, including a hard stage incentive reward function and a soft stage incentive reward function. Extensive experiments show that the soft stage incentive reward function is able to improve the convergence rate by up to 46.9% with the state-of-the-art DRL methods. The percentage increase in the convergence mean reward was 4.4-15.5% and the percentage decreases with respect to standard deviation were 21.9-63.2%. In the evaluation experiments, the success rate of trajectory planning for a robot manipulator reached 99.6%.

## I. INTRODUCTION

In order to use a robot manipulator to complete a task, it is essential to realize trajectory planning of the robot manipulator. The results of trajectory planning directly determine the quality index of the task carried out by the robot manipulator. Traditional trajectory planning for robot manipulators mainly includes artificial potential field methods [1-3] and polynomial interpolation [4-6]. These methods have low intelligence, poor dynamic planning, and no self-learning ability. In recent years, deep reinforcement learning (DRL) has been applied to trajectory planning of robot manipulators [7-10]. This method can allow a robot manipulator to learn autonomously and plan an optimal path in a complex and random environment.

Katyal [7], using only the original image of the working environment as the state space, demonstrated robust and direct mapping from the image to the action domain by exploiting simulation for learning based on DRL. The method is robust to environmental changes. It can learn a manipulation policy, which authors show takes the first steps toward generalizing to changes in the environment and can scale to new manipulators. But because the state space is an image, dimensionality disaster is likely to occur. Kamali [8] proposed a Dynamic-goal Deep Reinforcement Learning method to address the problem of robot arm motion planning in telemanipulation applications. This method intuitively maps human hand motions to a robot arm in real time while avoiding collisions, joint limits, and singularities. Li [9] presented an improved deep deterministic policy gradient algorithm to complete the task of trajectory planning based on a six-DOF arm robot. Wen [10] proposed to use a Deep Reinforcement Learning method to plan the trajectory of a robot arm to realize obstacle avoidance. In their approach, the rewards are designed to overcome the difficulty in the convergence of multiple rewards, especially when rewards are antagonistic to each other.

As shown in Fig. 1, the three components of DRL are environment, agent, and reward function. The agent in DRL performs exploration to identify the possible actions. The robot manipulator executes the action in the environment and feeds back the reward value to the agent according to the defined reward function. Through the iterative update method, the agent learns better strategies of trajectory planning.

---

This work was supported by National Natural Science Foundation of China (No.91748106) and Hubei Province Natural Science Foundation of China (No. 2019CFB526). *Corresponding author, Email: m201972630@hust.edu.cn.

Gang Peng, PhD, Assoc. Prof, IEEE Member, Email: penggang@hust.edu.cn;

Jin Yang (Co-First Author) Master graduate student, IEEE Student Member, Email:m201972630@hust.edu.cn;

Li Xinde, PhD, Prof, IEEE senior member; Email: xindeli@seu.edu.cn;

Mohammad Omar Khyam, Email: m.khyam@cqu.edu.au.

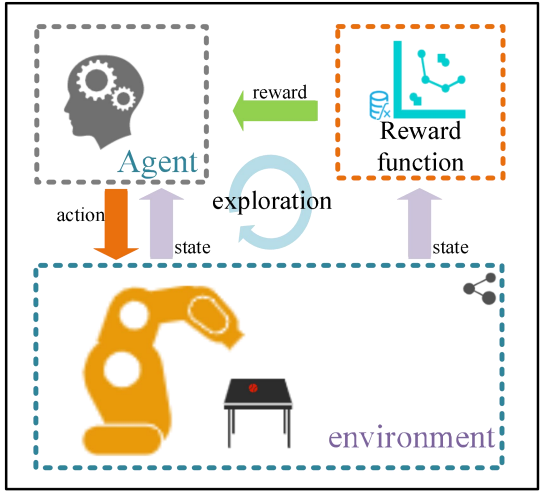

**FIGURE 1**. DRL framework.

In the history of the development of DRL, a typical method is the Deep Q-learning Network (DQN) [11-12]. However, its spaces of output action are discrete, so it is difficult to apply to continuous action spaces such as the trajectory planning of robot manipulators. Subsequently, Deep Deterministic Policy Gradient (DDPG) [13] based on the Actor-Critic (AC) architecture, Asynchronous Advantage Actor-Critic (A3C) [14], Proximal Policy Optimization (PPO) [15], and Soft Actor-Critic (SAC) [16] were proposed one by one [13-16].

However, there are still problems in DRL methods, such as randomness and blindness. The key to these problems is the reward function, which is an important part of DRL, but it can lead to a large amount of useless exploration and thus decrease the efficiency of the algorithm [17-19]. To solve the problem, we present a stage incentive mechanism based on human behavior cognition for robot trajectory planning in DRL. The primary contributions of this paper are summarized as follows:

1) Combining the characteristics of trajectory planning and work environment, three brand-new dense reward functions are proposed. Dense reward functions provide non-zero rewards, and they differ from the sparse reward function in that they provide more information after each action, which can reduce invalid and blind exploration of DRL during trajectory planning for the robot manipulator.

2) A posture reward function and a stride reward function are proposed. The posture reward function includes a position reward function and a direction reward function: the position reward function is composed of the task status item (whether or not the task is completed) and the distance guide item (the Euclidean distance between the end of the robot manipulator and the random target), and the direction reward function is modeled by the angle between the expected direction vector and the actual direction vector. The stride reward function includes a position reward and a movement distance reward. The position reward is the same as that mentioned above, and the movement distance reward is composed of the average movement distance of each joint of the robot manipulator. Together, the posture reward function and the stride reward function can make the robot manipulator explore more efficiently under reasonable constraints in position, direction, and movement distance, and reduce invalid and blind exploration.

3) In order to further improve learning efficiency, we are inspired by the cognitive process of human behavior and propose a stage incentive mechanism. The hard stage incentive mechanism is established by combining the posture reward function and stride reward function. To improve its potential stability hazards, a soft stage incentive mechanism is further proposed. With this innovative structure, we have increased the expected return obtained by the algorithm while ensuring the stability of the algorithm, which has improved the overall efficiency of the algorithm.

The rest of this paper is organized as follows. The structures of the posture reward function and stride reward function are presented in Section II and Section III. In Section IV, the stage incentive mechanisms are introduced, including the hard stage incentive reward function and the soft stage incentive reward function. The implementation of the reward functions is illustrated in Section V, mainly how to implement the proposed reward functions in the current mainstream DRL methods. Then, experimental results are demonstrated and discussed in Section VI. Finally, the conclusions are drawn in Section VII.

## II. POSTURE REWARD FUNCTION

For DRL-based methods, the robot manipulator performs a great deal of ineffective exploration in a complex random environment, which is the main reason for reducing the efficiency of the algorithm. A posture reward function restricts the relative position and relative direction of the endpoint of the robot manipulator and target reasonably by using a position reward function and a direction reward function, respectively. Therefore, a posture reward function can make the algorithm generate more reasonable actions to be executed by the robot manipulator and thereby improve the efficiency of the algorithm.

### A. POSITION REWARD FUNCTION

In a random environment, the Euclidean distance between the end of the robot manipulator and target can be used to reflect the current state of the robot. The position reward function is designed in this paper consists of two items, the task status and the distance guide. The task status item reflects the result of the trajectory planning, that is, whether the robot manipulator reaches the position of the target that appears in space randomly. The purpose of the distance guide item is to motivate the robot manipulator to approach the target point quickly.

*Distance guide item*: In order to motivate the robot manipulator to approach the target point T quickly, the distance guide item is represented by the Euclidean distance $D_{PT}$, between the end of the robotic arm $P$ and the target $T$.

*Task status item:* The task status item is modeled by $D_{PT}$. The smaller the $D_{PT}$, the more likely it is that the robot manipulator will reach the target. The task status item is represented by parameters $J_{reach}$: as shown in Eq. (1).

$$J_{reach} = \begin{cases} 0, D_{PT} > \beta \\ 1, D_{PT} < \beta \end{cases}, \quad (1)$$

where $\beta$ is adjustable according to the actual requirements of the environment, the value of $\beta$ is set to 0.01 in this paper.

By combining the task status item and distance guide item, the position reward function is designed as shown in Eq. (2).

$$R_{position}(D_{PT}) = J_{reach} - D_{PT}. \quad (2)$$

### B. DIRECTION REWARD FUNCTION

On the basis of the guidance of the position reward function, by adding a direction guide, the robot manipulator can obtain more information and reach the target faster.

The direction reward function is modeled by the relationship between two vectors in three-dimensional space: the expected direction and the actual direction of the end of the robot manipulator. As shown in Fig. 2, $PT$ is the expected motion direction, which is represented by $\overrightarrow{V_{PT}}$, and $PP'$ is the actual motion direction, which is represented by $\overrightarrow{V_{PP'}}$. The arithmetic expressions of $\overrightarrow{V_{PT}}$ and $\overrightarrow{V_{PP'}}$ are formulated in Eq. (3) and Eq. (4) as follows:

$$\overrightarrow{V_{PT}} = \langle (T_x - P_x), (T_y - P_y), (T_z - P_z) \rangle, \quad (3)$$

$$\begin{cases} \overrightarrow{V_{PP'}} = \langle (P'_x/temp), (P'_y/temp), (P'_z/temp) \rangle \\ temp = \sin(\cos^{-1}(P'_w)) \end{cases} \quad (4)$$

where $T_x, T_y, T_z$ are the coordinates of the target, $P_x, P_y, P_z$ are the coordinates of the end of the robot manipulator in the current state, $P'_x, P'_y, P'_z, P'_w$ is the quaternion of the end of the robot manipulator in the current state, and $\varphi$ represents the angle between $\overrightarrow{V_{PT}}$ and $\overrightarrow{V_{PP'}}$, which is applied to measure the deviation between the motion vector planned by the algorithm and the expected motion vector. The smaller the $\varphi$, the lower the deviation. The arithmetic expressions of $\varphi$ is formulated in Eq. (5).

$$\begin{cases} \varphi = \left| \cos^{-1} \frac{\overrightarrow{V_{PT}} \cdot \overrightarrow{V_{PP'}}}{\sqrt{(\overrightarrow{V_{PT}} \cdot \overrightarrow{V_{PT}}) \times (\overrightarrow{V_{PP'}} \cdot \overrightarrow{V_{PP'}})}} \right| \\ \varphi \in [0, \pi] \end{cases} \quad (5)$$

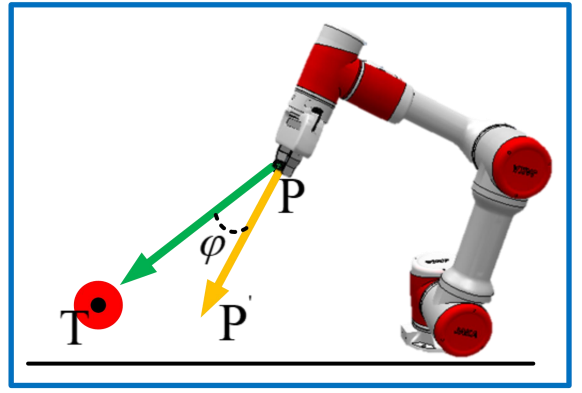

**FIGURE 2.** Scheme of the direction reward function.

The direction reward function designed in this paper is shown in Eq. (6).

$$R_{direction}(\varphi) = \lfloor \varphi \rfloor_* / 2\pi, \quad (6)$$

where $\lfloor \varphi \rfloor_*$ represents an operation in which the value of the function is output normally when the calculation result in $\lfloor \cdot \rfloor_*$ is less than $\pi/2$; otherwise, the result is $\pi - \varphi$.

### C. MODELING OF THE POSTURE REWARD FUNCTION

In the process of trajectory planning of the robot manipulator, if only the position reward function or only the direction reward function is used, the performance of the algorithm will be poor. Instead, the position reward function and direction reward function can be combined to form a posture reward function $R_{posture}$, as shown in Eq. (7).

$$R_{posture}(D_{PT}, \varphi) = R_{position}(D_{PT}) - R_{direction}(\varphi). \quad (7)$$

## III. STRIDE REWARD FUNCTION

The purpose of the stride reward function is to not only enable the robot manipulator to reach the target accurately but also make the movement distance of the robot manipulator is as small as possible for the optimal trajectory we expect. This allows the manipulator to reach the target quickly and reduces the energy consumption during the operation of the robot manipulator. The stride reward function is modeled by the position reward function and the movement distance reward function. The position reward function remains the same as in section II-A.

### A. MOVEMENT DISTANCE REWARD FUNCTION

In this article, we take the average movement distance of each joint as a constraint condition while the robot manipulator is running and model the movement distance reward function. It is difficult to obtain the movement distance of each joint directly during the operation of the robot manipulator. Therefore, we start from the speed of each joint of the robot manipulator to calculate the distance of each joint. We define the joint velocity vector of the robot manipulator as Eq. (8).

$$\vec{V} = [v_1, v_2, v_3, \ldots, v_N], N = number\ of\ joints. \quad (8)$$

The movement distance reward function $R_{move}$ is shown in Eq. (9).

$$R_{move}(\vec{V}) = \Delta t * (\vec{V} \cdot \vec{V})/N, \quad (9)$$

where $\Delta t$ represents the working frequency of the robot manipulator, that is, the robot manipulator runs according to the speed command every time at $\Delta t$. N is the number of joints of the robot manipulator. In this paper, we set the $\Delta t$ as 0.05, and $N$ as 6.

### B. MODELING OF THE STRIDE REWARD FUNCTION

The stride reward function proposed in this paper is a combination of the position reward function and the movement distance reward function. We use the position and the movement distance of each joint of the robot manipulator as constraints to promote the policy of the trajectory planning learned by the algorithm, which can ensure the target is reached and the movement distance of each joint of the robot manipulator is reduced.

The stride reward function designed in this paper is shown in Eq. (10).

$$R_{stride}(D_{PT},\vec{V}) = R_{position}(D_{PT}) - R_{move}(\vec{V}). \quad (10)$$

## IV. STAGE INCENTIVE MECHANISM

The stride reward function will restrict the motion of the robot manipulator. Actually, at the beginning of the task, we don't hope the robot was restricted, we hope the robot manipulator can move boldly to approach the target at this time. Therefore, we proposed a stage incentive mechanism.

### A. HARD STAGE INCENTIVE REWARD FUNCTION

We use an adjustable coefficient $\gamma$ to achieve the different reward functions at different stages of the task during the operation of the robot manipulator. The mechanism of the hard stage incentive divides the task of trajectory planning into two stages, including the fast approach area and the slow adjustable area, as shown in Fig. 3. In the fast approach area, the posture reward function is used to prompt the robot manipulator to approach the target quickly. In the slow adjustable area, the stride reward function is used as an incentive mechanism.

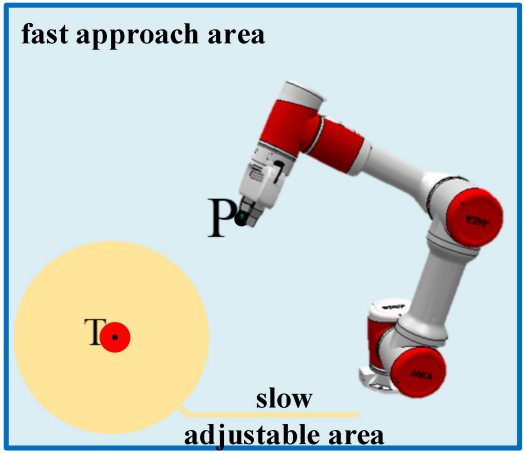

**FIGURE 3**. Scheme of the hard stage incentive reward function.

In this paper, we use $D_{PT} = 0.5$ as the boundary to divide the fast approach area and the slow adjustable area. The relationship between the adjustable coefficient $\gamma$ and the motion area of the robot manipulator is shown in Fig. 4.

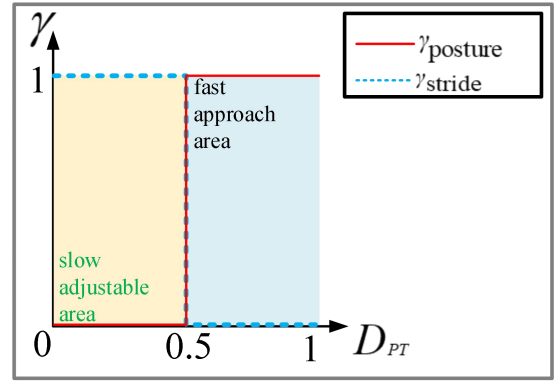

**FIGURE 4**. Diagram of the adjustable coefficient $\gamma$.

The value of $\gamma$ can be calculated by Eq. (11).

$$\gamma = \begin{cases} \begin{bmatrix} \gamma_{posture} = 1 \\ \gamma_{stride} = 0 \end{bmatrix}^T, P \in fast\ approach\ area \\ \begin{bmatrix} \gamma_{posture} = 0 \\ \gamma_{stride} = 1 \end{bmatrix}^T, P \in slow\ adjustable\ area \end{cases} \quad (11)$$

The mechanism of the hard stage incentive reward function $R_{HAR}$ we proposed is shown in Eq. (12):

$$R_{HAR} = \gamma \left[ R_{posture}(D_{PT}, \varphi)\ R_{stride}(D_{PT}, \vec{V}) \right]^T$$
$$= [\gamma_{posture}, \gamma_{stride}] \left[ R_{posture}(D_{PT}, \varphi), R_{stride}(D_{PT}, \vec{V}) \right]^T \quad (12)$$

### B. SOFT STAGE INCENTIVE REWARD FUNCTION

Although the hard stage incentive reward function achieved good results in experiments, we found that it has potential stability problems. That is, the adjustment process is rough, as it is easy to cause fluctuation of the reward curve when changing the reward function, which results in unstable factors for the algorithm. The switching process of the hard stage incentive reward function is similar to the bang-bang control in the classic control, and the method is bound to affect the stability of the algorithm. To solve this problem, we proposed the soft stage incentive reward.

In this paper, the weight coefficient $\alpha = [\alpha_1\ \alpha_2]$ is introduced to model a soft stage incentive reward function, as shown in Eq. (13) and Eq. (14):

$$\alpha_1 = f(D_{PT}) = 1 - \lfloor D_{PT} \rfloor_{-}^{\sigma_1}, \quad (13)$$

$$\alpha_2 = f(D_{PT}) = \lfloor D_{PT} \rfloor_{-}^{\sigma_2}, \quad (14)$$

where $\lfloor \cdot \rfloor_{-}$ represents an operation constraining the value of $D_{PT}$ in the range $[0,1]$, and $\sigma_1$ and $\sigma_2$ can be adjusted according to the actual situation of the task. In this paper, we set $\sigma_1 = \sigma_2 = 1$ according to experimental experience.

The soft stage incentive reward function, which adjusts the proportions of different reward functions through α, is defined as Eq. (15).

$$R_{SAR} = \alpha_1 R_{stride}(D_{PT}, \vec{V}) + \alpha_2 R_{posture}(D_{PT}, \varphi). \quad (15)$$

The soft stage incentive reward function does not need to divide the working space of the robot manipulator. According to the real-time change of the weight coefficient α, the reward function is adjusted dynamically and continuously.

## V. IMPLEMENTATION OF THE REWARD FUNCTION

As shown in Fig. 5, the learning process of the robot manipulator mainly consists of four stages: initialization, action generation, reward calculation and network training. The overall process is summarized in Algorithm 1.

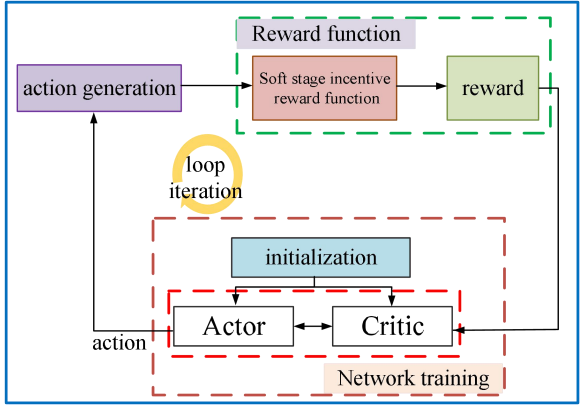

**FIGURE 5.** Diagram of the training process for DRL with an AC frame.

---

**Algorithm 1** Trajectory planning algorithm with the soft stage incentive reward function

**Input:** Environment state space S.

**Output:** Action a

1: Initialize Actor Network $\mu(S|\theta_\mu)$ and Critic Network $Q(S|\theta_Q)$

2: **for episode = 1 to M do**

3:    **for t = 1 to T do**

4:       $a_t \leftarrow \mu(S|\theta_\mu)$

5:       $R_{SAR} \leftarrow F(s,a)$

6:       reward = $R_{SAR}$

7:       Update weight of Actor Network $\theta_\mu$

8:       Update weight of Critic Network $\theta_Q$

9:    **end for**

10: **end for**

---

## VI. EXPERIMENTAL RESULTS AND DISCUSSIONS

In the experiment, we set 10000 episodes, and each episode has 50 steps. We used four indicators to evaluate the performance of our method: (1) convergence rate (expressed by the episode $E_{start}$ ($0 < E_{start} < 10000$) when the algorithm starts to converge), (2) the reward of each episode $R_{episode}$ is shown in Eq. (16) (during the experiment, we set any time step N in an episode to complete the trajectory planning and to immediately stop the current episode and enter the next episode; therefore, the total reward was calculated within the time step before completing the trajectory planning in each episode.), (3) the average number of steps to complete the task (it is impossible for the robot manipulator to complete the task of trajectory planning in one step), and (4) standard deviation $V_{STDEV}$ is shown in Eq. (17), is used to judge the stability and robustness of the algorithm.

$$R_{episode} = \sum_{s=1}^{N} Reward \ (1 \leq N \leq 50) \quad (16)$$

$$\begin{cases} \bar{R} = \frac{1}{(10000 - E_{start} + 1)} \sum_{i=E_{start}}^{i=10000} R_{episode} \\ V_{STDEV} = \sqrt{\frac{\sum_{i=E_{start}}^{n} \left(R_{episode}^i - \bar{R}\right)^2}{n-1}} \ (n = 10000) \end{cases} \quad (17)$$

Simulation experiments were conducted in V-REP [20], [21]. A random environment was initialized as shown in Fig. 6. The red ball is the target that randomly appears in the workspace. An additional reward of +20 was given after each task is successful.

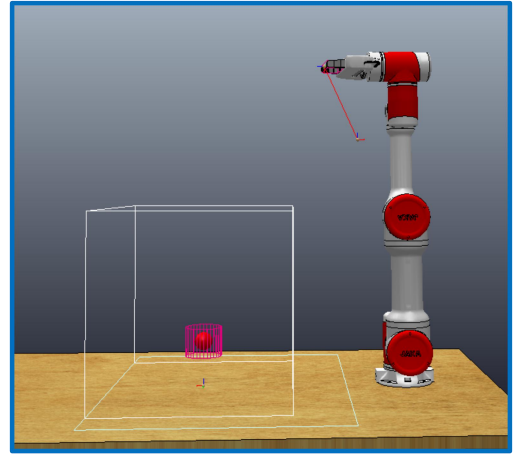

**FIGURE 6.** Simulation environments for the robot manipulator.

### A. POSTURE AND STRIDE REWARD FUNCTION

In this section, three types of reward functions, basic (a sparse reward function as shown in Eq. (18)), posture, and stride, were applied to two DRL methods. (According to experimental verification, DDPG and SAC could not converge even after a long training period based on a sparse reward function, so we do not discuss these scenarios.) During the experiments, we initialized the same working environment 20 times.

$$R_{basic} = \begin{cases} 1 & task\ is\ done \\ 0 & task\ not\ done \end{cases} \quad (18)$$

After all the methods converged, we calculated the convergence rate, the mean reward of each episode, the average steps to complete the task, and the standard deviation of the latter, as summarized in Table 1. The changing process of the reward and the average steps to complete the task in the training for each method are displayed in Fig. 7, and the changing process in the evaluation is shown in Fig. 8.

From Table 1, we can see that SAC converged faster in general. The convergence rate of SAC with posture reward function was 10.8% faster than that of DDPG with the posture reward function, and the convergence rate of SAC with the stride reward function was 17.9% faster than that of DDPG with the stride reward function. Compared with the stride reward function, the convergence rate of the posture reward function increased by 22.6-26.4%. However, the standard deviation of the stride reward function was 23.7-46.4% lower than that if the posture reward function, and the mean reward of the stride reward function was 3.2% higher than that of the posture reward function in DDPG.

As shown in Fig. 7, during the training, the posture reward function fluctuated to a large extent after convergence; in contrast, the stride reward function converged slowly, fluctuated greatly before convergence, and was stable after convergence. The reason is simple, posture reward function guides the robot manipulator closer to the target with distance and direction constraints, which is more oriented to complete the task. However, its stability is poor, and slight interference can make the manipulator move away from the target quickly, resulting in mission failure. The stride reward function takes the distance and the movement distance of each joint of the robot manipulator as constraints to guide the robot manipulator to approach the target. Hence, the manipulator will not suddenly move significantly. Compared to the posture reward function, the stride reward function is more cautious. The fluctuations in training are caused by the agent's exploration. Under the dual effects of exploration and the stride reward function's own characteristics, the robot manipulator cannot reach the target quickly.

During the evaluation, we conducted 500 random trials, used the trained model to realize the trajectory planning with the robot manipulator, and calculated its success rate. As shown in Table 1, the success rates of DDPG and SAC based on the posture reward function were 90.4% and 89.2%, respectively, and the success rates of DDPG and SAC based on the stride reward function were 88.6% and 90.8%, respectively.

Generally speaking, the posture reward function completes trajectory planning with fewer average steps, and the stride reward function obtains more rewards. In terms of improving the convergence rate, the posture reward function is more advantageous. The stride reward function plays a more important role in improving algorithm stability.

### B. HARD STAGE INCENTIVE REWARD FUNCTION

It can be seen from the above experiments that in a complex environment, the dense reward function will achieve better results, but there are still some defects. In this section, we applied our hard stage incentive reward function (hereafter referred to as HAR) to SAC and DDPG, and the convergence results are shown in Table 1. In the training, the changing process of the reward and the average steps to complete the task in the training for each method are shown in Fig. 7, and in the evaluation, the changing process for each method is visualized in Fig. 8.

As shown in Table 1, in the process of training, for robustness, the standard deviation decreased by about 42.6% with HAR compared to the posture reward function. The convergence rate of the HAR reward function was about 20.4% faster than that of the stride reward function, but it was slower than the posture reward function. When the mechanism of the hard stage incentive adjusted the reward function used in different stages, the switch-type adjustment method was adopted without a smooth transition process. This is also one of the reasons why the SAC with HAR fluctuated greatly in the training, as shown in Fig. 7(a), although its performance was not obvious in the DDPG.

In the evaluation, SAC and DDPG were greatly improved in both convergence performance and robustness when HAR was used. The reward increased to 24.7%, the average steps decreased by 16.7-27.2%, and the success rate for trajectory planning increased by 2.4-5.2%.

### C. SOFT STAGE INCENTIVE REWARD FUNCTION

Although HAR is able to offer improvements in both the training and evaluation processes, its learning efficiency and robustness still need to be improved.

**TABLE 1.** Results with the stage incentive reward function.

| Method | Reward Function | Train | | | | Evaluation | | |
|---|---|---|---|---|---|---|---|---|
| | | Episode | Reward | Step | Standard deviation | Reward | Step | Success rate |
| SAC | Basic | ---- | ---- | ---- | ---- | ---- | ---- | ---- |
| | Posture | 5244 | 9.956±1.352 | 15 | 18.161±1.430 | 13.902 | 10 | 89.2% |
| | Stride | 6773 | 16.125±0.833 | 12 | 9.728±1.049 | 16.584 | 12 | 90.8% |
| | HAR | 5369 | 15.366±0.403 | 12 | 10.358±0.348 | 17.349 | 10 | 93.2% |
| | SAR | 3593 | 16.830±0.255 | 10 | 7.594±0.222 | 17.969 | 7 | 97.0% |
| DDPG | Basic | ---- | ---- | ---- | ---- | ---- | ---- | ---- |
| | Posture | 5879 | 15.644±0.518 | 9 | 11.625±3.073 | 14.871 | 11 | 90.4% |
| | Stride | 7988 | 16.142±0.827 | 11 | 8.872±2.505 | 16.163 | 11 | 88.6% |
| | HAR | 6385 | 17.266±0.276 | 9 | 6.688±0.549 | 17.361 | 8 | 93.8% |
| | SAR | 4781 | 18.076±0.166 | 8 | 4.283±0.305 | 18.96 | 5 | 99.6% |

In the last set of experiments, the soft stage incentive reward function (hereafter referred to as SAR) was used. As shown in Table 1, the results of SAR were superior in all cases. In the training, the changing process of the reward and the average steps to complete the task in the training for each method are shown in Fig. 7, and the changing process in the evaluation is shown in Fig. 8.

In the training, compared with the above three reward functions, the convergence rate was accelerated by 18.7-40.1% in DDPG and by 31.4-46.9% in SAC. For the convergent mean reward, the promotion was between 4.6-15.5% in DDPG and 4.4-9.5% in SAC. The performance of robustness was also excellent; the standard deviation decreased by 35.9-63.2% in DDPG and by 21.9-26.7% in SAC. This shows SAR has a great convergence rate, stability, and robustness. Why does it work so well? One reason is that it combines the advantages of the posture reward function and the stride reward function to ensure fast and stable convergence in the early stages of exploration. Another reason is that it solves the switch adjustment mode of HAR by smoothing the transition between different reward functions in different stages.

The convergence rate of SAR in DDPG was slower than that of SAR in SAC, but other indicators were better in DDPG than in SAC. With the use of the model obtained by the DDPG with SAR for evaluation, the success rate of the trajectory planning reached 99.6%. The average number of steps to complete the trajectory planning was 5. From Fig. 8, compared with the other three reward functions, we can observe that SAR needs fewer steps to realize trajectory planning of the robot manipulator, obtains more rewards, and is more stable.

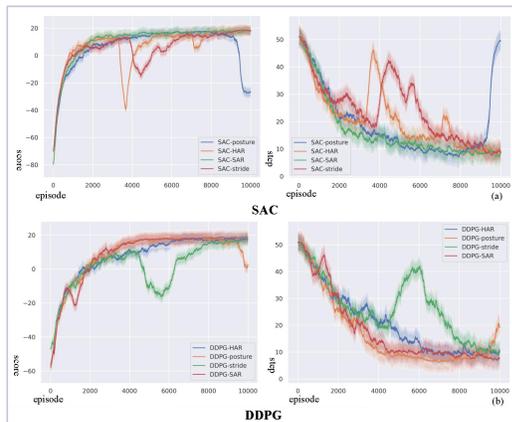

**FIGURE 7.** Diagram of the convergence process with the posture and stride reward function in the training.

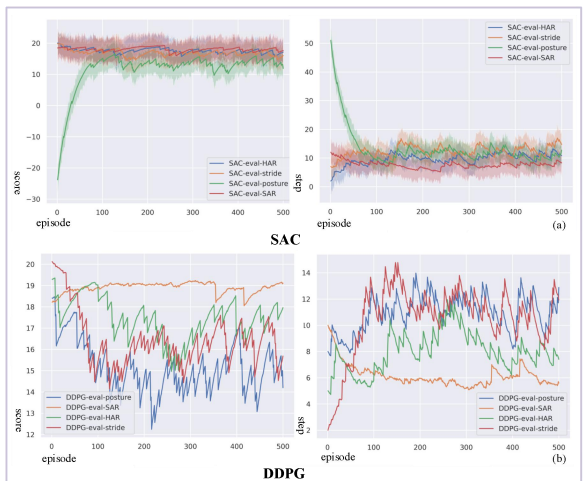

**FIGURE 8.** Diagram of the process with the posture and stride reward function in the evaluation.

## VII. CONCLUSIONS

To deal with the inefficiency, instability, and blindness of DRL-based methods in trajectory planning, this paper proposes three dense reward functions: the posture reward function, stride reward function, and the mechanism of stage incentive. The posture reward function can reduce the blindness of exploration to accelerate the learning process, and the stride reward function can make the learning process more stable. However, the soft stage incentive reward function exhibits the advantages of both, offering faster convergence, higher stability, and greater robustness. The experimental results show that state-of-the-art DRL methods using the proposed reward functions will have the best performance.

In the future, we will further explore the mechanism of reward shaping, and we plan to apply DRL methods to other complex tasks with robot manipulators.

## REFERENCES

[1] N. Zhang, Y. Zhang, C. Ma and B. Wang, "Path planning of six-DOF serial robots based on improved artificial potential field method," 2017 IEEE International Conference on Robotics and Biomimetics (ROBIO), Macau, 2017, pp. 617-621.

[2] H. Lin and M. Hsieh, "Robotic Arm Path Planning Based on Three-Dimensional Artificial Potential Field," 2018 18th International Conference on Control, Automation and Systems (ICCAS), Daegwallyeong, 2018, pp. 740-745.

[3] S. N. Gai, R. Sun, S. J. Chen and S. Ji, "6-DOF Robotic Obstacle Avoidance Path Planning Based on Artificial Potential Field Method," 2019 16th International Conference on Ubiquitous Robots (UR), Jeju, Korea (South), 2019, pp. 165-168.

[4] S. Fang, X. Ma, Y. Zhao, Q. Zhang and Y. Li, "Trajectory Planning for Seven-DOF Robotic Arm Based on Quintic Polynormial," 2019 11th International Conference on Intelligent Human-Machine Systems and Cybernetics (IHMSC), Hangzhou, China, 2019, pp. 198-201.

[5] S. Fang, X. Ma, J. Qu, S. Zhang, N. Lu, X. Zhao (2020) Trajectory Planning for Seven-DOF Robotic Arm Based on Seventh Degree Polynomial. In: Y. Jia, J. Du, W. Zhang (eds) Proceedings of 2019 Chinese Intelligent Systems Conference. CISC 2019. Lecture Notes in Electrical Engineering, vol 593. Springer, Singapore.

[6] Sadiq A T , Raheem F A , Abbas N A F . Optimal Trajectory Planning of 2-DOF Robot Arm Using the Integration of PSO Based on D* Algorithm and Cubic Polynomial Equation[C]// The First International Conference for Engineering Researches - March 2017. 2017.

[7] K. Katyal, I. Wang and P. Burlina, "Leveraging Deep Reinforcement Learning for Reaching Robotic Tasks," 2017

IEEE Conference on Computer Vision and Pattern Recognition Workshops (CVPRW), Honolulu, HI, 2017, pp. 490-491.

[8] K. Kamali, I. A. Bonev and C. Desrosiers, "Real-time Motion Planning for Robotic Teleoperation Using Dynamic-goal Deep Reinforcement Learning," 2020 17th Conference on Computer and Robot Vision (CRV), Ottawa, ON, Canada, 2020, pp. 182-189.

[9] Z. Li, H. Ma, Y. Ding, C. Wang and Y. Jin, "Motion Planning of Six-DOF Arm Robot Based on Improved DDPG Algorithm," 2020 39th Chinese Control Conference (CCC), Shenyang, China, 2020, pp. 3954-3959.

[10] S. Wen, J. Chen, S. Wang, H. Zhang and X. Hu, "Path Planning of Humanoid Arm Based on Deep Deterministic Policy Gradient," 2018 IEEE International Conference on Robotics and Biomimetics (ROBIO), Kuala Lumpur, Malaysia, 2018, pp. 1755-1760.

[11] V. Mnih, K. Kavukcuoglu, D. Silver, et al. Human-level control through deep reinforcement learning. Nature 518, 529–533 (2015).

[12] T. D. Le, A. T. Le and D. T. Nguyen, "Model-based Q-learning for humanoid robots," 2017 18th International Conference on Advanced Robotics (ICAR), Hong Kong, 2017, pp. 608-613.

[13] T.P. Lillicrap, J.J. Hunt, A. Pritzel, N. Heess, T. Erez, Y. Tassa, D. Silver, and D. Wierstra, "Continuous control with deep reinforcement learning," arXiv preprint arXiv:1509.02971, 2015.

[14] V. Mnih, A.P. Badia, M. Mirza, A. Graves, T. Lillicrap, T. Harley, and K. Kavukcuoglu, "Asynchronous methods for deep reinforcement learning." in Proc. ICML, New York, USA, Jun. 2016, pp.1928–1937.

[15] J. Schulman, F. Wolski, P. Dhariwal, A. Radford, O. Klimov, "Proximal Policy Optimization Algorithms." arXiv preprint arXiv:1707.06347, 2017.

[16] T. Haarnoja, A. Zhou, P. Abbeel, S. Levine, "Soft Actor-Critic: Off-Policy Maximum Entropy Deep Reinforcement Learning with a Stochastic Actor", arXiv preprint arXiv:1801.01290, 2018.

[17] J. Xie, Z. Shao, Y. Li, Y. Guan and J. Tan, "Deep Reinforcement Learning with Optimized Reward Functions for Robotic Trajectory Planning," in IEEE Access, vol. 7, pp. 105669-105679, 2019.

[18] S. Jang and M. Han, "Combining Reward Shaping and Curriculum Learning for Training Agents with High Dimensional Continuous Action Spaces," 2018 International Conference on Information and Communication Technology Convergence (ICTC), Jeju, 2018, pp. 1391-1393.

[19] A. Hussein, E. Elyan, M. M. Gaber and C. Jayne, "Deep reward shaping from demonstrations," 2017 International Joint Conference on Neural Networks (IJCNN), Anchorage, AK, 2017, pp. 510-517.

[20] E. Rohmer, S.P. Singh, and M. Freese, "V-REP: A versatile and scalable robot simulation framework," in Proc. IROS, Tokyo, Japan, Nov. 2013, pp.1321-1326.

[21] M. Freese, S.P. Singh, and F. Ozaki, "Virtual Robot Experimentation Platform V-REP: A Versatile 3D Robot Simulator," in Proc. ICS, Darmstadt, Germany, Nov. 2010, pp.536-541.